\documentclass{article}

\PassOptionsToPackage{numbers, compress}{natbib}
\usepackage[dblblindworkshop, final]{main}

\usepackage[utf8]{inputenc} % allow utf-8 input
\usepackage[T1]{fontenc}    % use 8-bit T1 fonts
\usepackage{hyperref}       % hyperlinks
\usepackage{url}            % simple URL typesetting
\usepackage{booktabs}       % professional-quality tables
\usepackage{amsfonts}       % blackboard math symbols
\usepackage{nicefrac}       % compact symbols for 1/2, etc.
\usepackage{microtype}      % microtypography
\usepackage{xcolor}         % colors

\usepackage{algorithm}
\usepackage{algorithmic}
\usepackage{subfig}
\usepackage{graphicx}
\PassOptionsToPackage{numbers, compress}{natbib}

\newcommand{\INT}{{\tt{INTERACT\,}}}
\newcommand{\PXP}{{\tt{PXP\,}}}
\newcommand{\PEX}{{\tt{PEX\,}}}

\newcommand{\Ratify}{{\mathit{RATIFY}}}
\newcommand{\Revise}{{\mathit{REVISE}}}
\newcommand{\Refute}{{\mathit{REFUTE}}}
\newcommand{\Reject}{{\mathit{REJECT}}}

\newcommand{\Init}{{\mathit{INIT}}}
\newcommand{\Match}{{\mathtt{MATCH}}}

\newcommand{\Agree}{{\mathtt{AGREE}}}

\newcommand{\Interact}{{\mathtt{INTERACT}}}
\newcommand{\CallAgent}{{\mathtt{AGENT}}}
\newcommand{\AskAgent}{{\mathtt{ASK\_AGENT}}}
\newcommand{\UpdateContext}{{\mathtt{UPDATE\_CONTEXT}}}
\newcommand{\AssemblePrompt}{{\mathtt{ASSEMBLE\_PROMPT}}}
\newcommand*\samethanks[1][\value{footnote}]{\footnotemark[#1]}

\newtheorem{mydefinition}{Definition}

\title{
Multi-Turn Human–LLM Interaction Through the Lens of a Two-Way Intelligibility Protocol}

\author{%
Harshvardhan Mestha \\
  Dept. of Electrical and Electronics Engineering \\
  BITS Pilani, K.K. Birla Goa Campus \\
  Zuarinagar 403726, Goa, India \\
  \texttt{f20220609@goa.bits-pilani.ac.in} \\
  \And
  Karan Bania\thanks{Work done while at BITS Pilani, K.K Birla Goa Campus} \\
  Machine Learning Department \\
  Carnegie Mellon University \\
  \texttt{kbania@cs.cmu.edu} \\
  \And
  Shreyas Vinaya Sathyanarayana\samethanks \\
  Mstack AI \\
  \texttt{shreyas.college@gmail.com} \\
  \And
  Sidong Liu \\
  Centre for Health Informatics \\
  Macquarie University, Sydney \\
  \texttt{sidong.liu@mq.edu.au} \\
  \And
  Ashwin Srinivasan \\
  Dept. of Computer Science \& Information Systems \\
  BITS Pilani, K.K. Birla Goa Campus \\
  Zuarinagar 403726, Goa, India \\
  \texttt{ashwin@goa.bits-pilani.ac.in} \\
}

\begin{document}

\maketitle

\begin{abstract}
Our interest is in the design of software systems involving a human-expert
interacting--using natural language--with a large language model (LLM) on data analysis tasks.
For complex problems, it is possible that LLMs can harness human expertise and creativity
to find solutions that were otherwise elusive.
On one level, this interaction takes place through
multiple turns of prompts from the human and responses from the LLM. Here we investigate
a more structured approach based on an abstract protocol described in~\cite{pxp}
for interaction between agents. The protocol
is motivated by a notion of ``two-way intelligibility'' and is
modelled by a pair of communicating finite-state machines.
We provide an implementation of the protocol,
and provide empirical evidence of using the implementation 
to mediate interactions between an LLM and
a human-agent in two areas of
scientific interest (radiology and drug design).
% Experiments conducted are in the form of
% (a) controlled experiments using a human-proxy
% (a database labelled with human-expertise: the proxy); and
% (b) uncontrolled experiments with human-subjects with different
% levels of expertise.
We conduct controlled experiments with a human proxy (a database), and uncontrolled experiments with human subjects.
The results provide evidence in support
of the protocol's capability of capturing one- and two-way intelligibility in
human-LLM interaction; and
 for the utility of two-way intelligibility in the design
of human-machine systems.
Our code is available at \href{https://github.com/karannb/interact}{\texttt{https://github.com/karannb/interact}}.

% in the manner proposed in~\cite{pxp}. The results from the uncontrolled
% results are consistent with those observed in controlled experiments, while
% providing interesting insights into the role of prior human-expertise. Together the
% results provide support for utility of two-way intelligibility in the design
% of human-machine systems for data analysis.
\end{abstract}

\section{Introduction}

For over four decades, researchers in human–machine systems have recognised the importance of making machine-generated predictions intelligible to humans. Michie \cite{michie:window82} was among the first to note potential mismatches between human and machine representations. He proposed a classification of machine learning (ML) systems into three types \cite{michie:ewsl88}: 
Weak ML systems are concerned only
with improving performance, given sample data. 
Strong ML systems improve performance, but are
also required to communicate what it has learned in some human-comprehensible form (Michie assumes this will be symbolic). 
Ultra-strong ML systems are Strong ML systems that can also teach the human to improve his or her performance.

This framework, intended for human-in-the-loop contexts, focuses on intelligibility, the ability of the system to explain the “what” and “why” within the human’s window of understanding rather than intelligence. The system’s internal model can use any representation, so long as communication is understandable. Michie’s scheme addresses one-way communication from machine to human, without metrics for intelligibility (later addressed in \cite{mugg:expl}) or provisions for human-to-machine communication, which is important in collaborative discovery. This categorisation has also recently informed a similar 3-way categorisation for the use of AI tools in scientific discovery \cite{krenn:nature2022}.

Despite growing interest in human–AI interaction \cite{carolin:2021:extendedAI,andres:2024:designforHAI,tsiakas:2024:unpackinghai,jing:2024:researchonHAI}, few models prioritise mutual intelligibility. Explainable AI (XAI) often produces explanations without tailoring them to the intended audience, undermining user confidence when advice is opaque and when the system cannot gauge the user’s comprehension. A recent proposal \cite{pxp} addresses this by defining “two-way intelligibility” as a property of communication protocols between agents, modelled as finite-state machines. Messages are tagged with one of four “R’s”: ratification, refutation, revision, or rejection, and intelligibility is assessed from the sequence of these tags. While \cite{pxp} specifies the protocol with proofs of correctness and termination, no implementation or empirical evidence was provided.

Given the capabilities of large language models (LLMs) to produce human-readable predictions and explanations when properly conditioned, they are promising candidates for such protocols. However, determining the right conditioning information is non-trivial. This paper addresses that gap by implementing the \cite{pxp} protocol restricted to interactions between an LLM-based generator-agent and a human tester-agent, providing a partial but practical realisation of two-way intelligibility in action.

\section{Related Work}
\textbf{Iterative Prompting}: Work has been done in prompting LLMs iteratively, by decomposing the problem into smaller sub-problems, or paths in a decision tree, such as Chain-of-Thought \cite{wei2023chainofthoughtpromptingelicitsreasoning,sun2024enhancingchainofthoughtspromptingiterative}, Tree-of-Thought \cite{yao2023treethoughtsdeliberateproblem}, etc. Our work is not a method to iteratively prompt an LLM; rather, it is a method to analyse and interpret the interactions between a Human and an LLM in a multi-turn fashion, using the notion of intelligibility. 

\textbf{Intelligibility}: Intelligibility is the ability to be understood or comprehended. Previous work has proposed using concepts in philosophy, psychology, and cognitive science \cite{miller2018explanationartificialintelligenceinsights, psychai2022} to gain insight into intelligible systems.  Our aim is to measure intelligibility and identify Strong/Ultra Strong ML systems as described by Michie~\cite{michie:ewsl88}, and see if the characteristics are shared across tasks. Our implementation allows us to explore interactions between not only Human-ML interactions but also interactions between two ML agents. We achieve this by restricting the Agents to Predict and Explain (PEX) as described in~\cite{pxp}. This constraint enables us to quantify intelligibility using message tags.

% The rest of the paper is organised as follows. We provide a brief
% summary of the main aspects \PXP protocol in Sec.~\ref{sec:spec}.
% A simple implementation  of the protocol
% suitable for an agent that uses an LLM is in Sec.~\ref{sec:impl}.
% Sec.~\ref{sec:expt} reports on experiments with using
% the protocol for tasks in computational drug-discovery and
% radiology report generation. Sec.~\ref{sec:concl} concludes the paper.

% \section{Related Work}
% \label{sec:relwork}
% \cite{chaves:2024:radiologyfoundation}(sent abt rad foundation model here)
% \cite{tang:2025:chemagent}(sent abt chemagent here)

\section{The Main Aspects of the PXP Protocol}
\label{sec:spec}

The \PXP protocol is an interaction model described in~\cite{pxp}.
It is motivated by the question of when an interaction between agents could be said to
be intelligible to either. Restricting attention to
specific kinds of agents--called \PEX agents--the paper is concerned
with when predictions and explanations for a data instance produced by one \PEX
agent are intelligible to another \PEX agent. The agents are modelled
as finite-state machines, and ``intelligibility'' is defined as a property
inferrable from the tagged messages exchanged between the finite-state machines.

An agent $a_n$ decides the tag of its message to be sent to
agent $a_m$ based on comparing its prediction $y_n$ and explanation
$e_n$ for a data instance $x$ to the 
prediction and explanation it received from $a_m$ for $x$ ($y_m$ and
$e_m$ respectively). The comparisons define the truth-value of guards
for transitions, which in turn define the message-tags. 
Informally, if $a_n$ agrees with $a_m$ on both
prediction and explanation, it sends a message with a $\Ratify$ tag. If $a_n$ disagrees
with either the prediction or explanation but not both, then it might change
(``revise'') its (internal) model. If it can revise its model, then its message has
a $\Revise$ tag; else it has a $\Refute$ tag. Finally, if $a_n$ disagrees with
both prediction and explanation, it sends a $\Reject$ tag. In all cases, the complete
message has the tag, along with $a_n$'s current prediction and explanation.
Intelligibility is then defined as properties inferable from observing
the message-tags exchanged. The important definitions in~\cite{pxp} are: %(in~\cite{pxp}, $a_m, m$ and $a_n, n$ are used interchangeably
%for readability):

% \begin{mydefinition}[One-Way Intelligibility]
\begin{mydefinition}[One-Way Intelligibility]
\label{def:1wayintell}
Let $S$ be a session between compatible
agents $a_m$ and $a_n$ using $\PXP$. Let $T_{mn}$ and $T_{nm}$ be the
sequences of message-tags sent in a session $S$ from $m$ to $n$ and from $n$ to $m$. 
We will say $S$ exhibits One-Way Intelligibility for $m$ iff
(a) $T_{mn}$ contains at least one element in
    $\{\Ratify,\Revise\}$; and
(b) there is no $\Reject$ in the sequence $T_{mn}$.  
Similarly for $n$.
\end{mydefinition}

% \noindent
Two-way intelligibility follows directly if a session $S$ is one-way intelligible
for $m$ and one-way intelligible for $n$. The authors in~\cite{pxp} then
use these definitions to suggest a Michie-style categorisation of
Strong and Ultra-Strong Intelligibility. Based on this suggestion, we
define the following:

\begin{mydefinition}[Strong and Ultra-Strong Intelligibility]
\label{def:strong}
Let $S$ be a session between compatible agents $a_m$ and $a_n$.
 Let $T_{mn}$ and $T_{nm}$ be the
sequences of message-tags sent in a session $S$ from $m$ to $n$ and from $n$ to $m$. 
We will say $S$ exhibits Strong Intelligibility for $m$ iff
every element of $T_{mn}$ is from $\{\Ratify,\Revise\}$.
We will say $S$ exhibits Ultra-Strong Intelligibility for $m$ iff
$S$ exhibits Strong Intelligibility for $m$ and there is
at least one element of $T_{mn}$ is a $\Revise$. Similarly for
$a_n$.
\end{mydefinition}
% \textbf{One-Way Intelligibility: }Let $S$ be a session between compatible
% agents $a_m$ and $a_n$ using $\PXP$. Let $T_{mn}$ and $T_{nm}$ be the
% sequences of message-tags sent in a session $S$ from $m$ to $n$ and from $n$ to $m$. We will say $S$ exhibits One-Way Intelligibility for $m$ iff
% (a) $T_{mn}$ contains at least one element in
%     $\{\Ratify,\Revise\}$; and
% (b) there is no $\Reject$ in the sequence $T_{mn}$.Similarly for $n$. Two-way intelligibility follows directly if a session $S$ is one-way intelligible
% for $m$ and one-way intelligible for $n$. The authors in~\cite{pxp} then
% use these definitions to suggest a Michie-style categorisation of
% Strong and Ultra-Strong Intelligibility. Based on this suggestion, we
% define:
% % \begin{mydefinition}[Strong and Ultra-Strong Intelligibility]
% \textbf{Strong and Ultra-Strong Intelligibility: }
% \label{def:strong}
% Let $S$ be a session between compatible agents $a_m$ and $a_n$.
%  Let $T_{mn}$ and $T_{nm}$ be the
% sequences of message-tags sent in a session $S$ from $m$ to $n$ and from $n$ to $m$. 
% We will say $S$ exhibits Strong Intelligibility for $m$ iff
% every element of $T_{mn}$ is from $\{\Ratify,\Revise\}$.
% We will say $S$ exhibits Ultra-Strong Intelligibility for $m$ iff
% $S$ exhibits Strong Intelligibility for $m$ and there is
% at least one element of $T_{mn}$ is a $\Revise$.
Similarly for
$a_n$.We now examine a simple implementation of \PXP for modelling interactions
in which $a_m$ and $a_n$ are agents that have access to predictions and
explanations originating from LLMs and human-expertise, respectively. For simplicity,
we refer to the former as the ``machine-agent'' and the latter as the ``human-agent'',
and the interaction between $a_m$ and $a_n$ as ``human--LLM interaction''.

\section{Modelling Human-LLM Interaction}\label{sec:impl}
The implementation described here is restricted to interaction
between a single human-agent and a single machine-agent.
The protocol in~\cite{pxp} is akin
to a plain-old-telephone-system (POTS), and it is sufficient
for our purposes to implement it as a rudimentary blackboard system
with a simple scheduler that alternates between the two
agents. The blackboard consists of 3 tables accessible to the agents.
The tables are:
(a) $Data$. This is a table consisting of $(s,x)$ pairs
    where $s$ is a session ID, and $x$ is a data instance; and
(b) $Message$. A table of 5-tuples $(s,j,\alpha,\mu,\beta)$
    where: $s$ is a session identifier, $j$ is a message-number,
    $\alpha$ is a sender-id, $\mu$ is a message and $\beta$ is a receiver-id; and
(c) $Context$. This a table consisting of 3-tuples $(s,j,c)$ where
    $s$ is a session-id, $j$ is the message number, and $c$ is some
    domain-specific context information. 
For simplicity, message-numbers will be assumed to be from the
set $\{0,1,2,\ldots\}$; $\alpha,\beta$ are from
$\{h,m\}$ where $h$ denoting ``human'' and $m$ denotes ``machine'';
and  messages $\mu$ are $(l,y,e)$ tuples, where $l$ is from
$\{\Ratify,\Refute,\Revise,\Reject\}$, and $y$ and $e$ are 
the prediction and explanation respectively.
We treat the blackboard as a relational database $\Delta$
consisting of the set of tables $\{D,M,C\}$. For presentability, $\Delta$ is denoted as a ``shared'' input in
the agent-functions below.

The procedure in Alg.~\ref{alg:hloop} implements the interaction.
The interaction is initiated by the machine, with its
prediction and explanation for a data instance (one-per-session),
and proceeds until
all data instances have been examined. 
% The procedure returns
% a summary of the interaction in the form of the contents of the
% common storage. This contains
% a record of the sessions in terms of the data provided (table $D$),
% and messages exchanged (table $M$).
% This information is mainly generated through the use of a function called
% $\AskAgent$, implemented as in Alg.~\ref{alg:askagent}. 
% These functions ask the corresponding
% machine and human-agents to provide: (a) an assessment of the prediction
% and explanation provided for the data instance; and (b) the agent's own
% prediction and explanation from the data instance. The assessment is a message-tag, obtained in the manner described in~\cite{pxp} and
% summarised in Sec.~\ref{sec:spec}, with the following 
% small modification:
The function $\AskAgent$ (Alg.~\ref{alg:askagent}) asks the corresponding agent to obtain the prediction and explanation from the corresponding agent. The assessment is a message tag which is obtained using the $\CallAgent$ (Alg.~\ref{alg:callagent}) procedure.
     We assume that both agents only send a $\Reject$ tag
        after the interaction has proceeded for some minimum number
        of messages. Until then, the machine's message will be tagged  either as
        $\Revise$ 
        % (if it is able to revise its output to match one or
        % both of the human-agent's prediction or explanation) 
        or $\Refute$. % otherwise.
        After the bound, the machine can send a message with a $\Reject$ tag.
        Similarly, the human-agent will send a $\Refute$ message to the machine until
        this bound is reached, after which the message tag can be $\Reject$.
        Since the message-length is bounded, it will not affect the termination
        properties of the bounded version of \PXP. 
The procedure for calling the human- or machine-agent  is in 
Alg.~\ref{alg:callagent}. Unsurprisingly, the
same procedure suffices for both kinds of agents, since \PXP is a symmetric protocol that
does not distinguish between agents (other than a special agent called
the oracle). Agent-specific details arise in
the $\Match$ and $\Agree$ relations, and in the question-answering
step (the $\AskAgent$ function), which is shown in Alg.~\ref{alg:askagent}.
The $\AssemblePrompt$ is domain-dependent, and not described algorithmically here.
Instead, we present it by example below. The $\Match$ and $\Agree$ functions
used are described in section, Sec.~\ref{sec:expt}. We next report results from experiments using this implementation.

\begin{algorithm}[!htb]
\small

\caption{The $\CallAgent$ Procedure.}
\label{alg:callagent}
    \textbf{Input}:
        $q$ an agent query;
        $\lambda$: an agent identifier;
        $s$: a session identifier;
        $j$: a message-number ($j >0$);
        $k$; message-number after with $\Reject$ tags can be sent $(k > 1)$;
        $\Delta$: a shared relational database;\\
    \textbf{Output}: $(l,y,e)$, where
        $l \in \{\Init, \Ratify, \Reject, \Revise, \Refute\}$,
        $y$ is a prediction, and 
        $e$ is an explanation
    \begin{algorithmic}[1]
        \STATE Let $\Delta = \{D,M,C\}$\;
        \STATE Let $(s,x) \in D$\;
        \STATE $C_0 ;= \emptyset$\;
        \STATE $(y,e):= \AskAgent(q,x,\lambda,C_{j-1})$\;
        \IF{$(j \geq 2)$}
            \STATE Let $(s,j-1,\alpha,(l',y',e'),\lambda) \in M$\;
            \IF{$(j=2)$}
                \STATE $y'' := y$\;
                \STATE $e'' := e$\;
            \ELSE
                \STATE Let $(s,j-2,\lambda,(l'',y'',e''),\alpha) \in M$\;
            \ENDIF
            \STATE $CatA := (\Match_\lambda(y',y'') \wedge \Agree_\lambda(e',e''))$\;
            \STATE $CatB := (\Match_\lambda(y',y'') \wedge \neg \Agree_\lambda(e',e''))$\;
            \STATE $CatC := (\neg\Match_\lambda(y',y'') \wedge \Agree_\lambda(e',e''))$\;
            \STATE $CatD := (\neg\Match_\lambda(y',y'') \wedge \neg\Agree_\lambda(e',e''))$\;
            \STATE $Changed := (\neg \Match_\lambda(y,y'') \vee \neg \Agree_\lambda(e,e''))$\;
            \IF{$(CatA)$}
                \STATE $l := \Ratify$\;
            \ELSIF{$(CatB$ or $CatC)$}
                \IF{$(Changed)$}
                    \STATE $l := \Revise$\;
                \ELSE
                    \STATE $l := \Refute$\;
                \ENDIF
            \ELSIF{$(CatD)$}
                \IF{$(j > k)$}
                    \STATE $l := \Reject$\;
                \ELSIF{$Changed$}
                    \STATE $l := \Revise$\;
                \ELSE
                    \STATE $l := \Refute$\;
                \ENDIF
            \ENDIF
        \ELSE
            \STATE $l := Init$\;
        \ENDIF
        \STATE $C_j := \UpdateContext((l,y,e), j, C_{j-1})$\;
        \STATE $C := C \cup \{(s,j,C_j)$\}\;
        \RETURN $(l,y,e)$
    \end{algorithmic}
\end{algorithm}

\begin{algorithm}[!htb]
% \small

\caption{The $\AskAgent$ Function.}
\label{alg:askagent}
    \textbf{Input}:
        $q$: an agent query;
        $x$: a data instance;
        $\lambda$: an agent identifier;
        $C$: prior information;\\
    \textbf{Output}:  
        $(y,e)$, where
        $y$ is a prediction, and 
        $e$ is an explanation
    \begin{algorithmic}[1]
        \IF{$(\lambda$ is an LLM$)$}
            \STATE $P := \AssemblePrompt(q,C)$
            \STATE $(y,e) := \lambda(x|P)$\;     // ask an LLM
        \ELSE
            \STATE $(y,e) = \lambda(x|q,C)$\;   // ask the other agent
        \ENDIF
        \RETURN $(y,e)$
    \end{algorithmic}
\end{algorithm}

\section{Experimental Evaluation}
\label{sec:expt}

\begin{algorithm}
    % \small
    \caption{The $\Interact$ Procedure.}
    \label{alg:hloop}
    \textbf{Input}:
        $X$: a set of data instances;
        $h$: a human-agent identifier;
        $m$: a machine-agent identifier;
        $q_h$: a query for the human-agent;
        $q_m$: a query for the machine-agent;
        $n$: an upper-bound on the total number of messages in an interaction $(n >0)$;
        $k$: message-number after which an agent can send $\Reject$ tags in messages $(k \geq 1$); \\
    \textbf{Output}: A relational database $\Delta = \{D,M,C\}$\;
    
    \begin{algorithmic}[1] %[1] enables line numbers
    \STATE $Left := X$\;
    \STATE $D = M = C := \emptyset$\;
    \STATE Let $\Delta = \{D, M, C\}$\;
    \STATE Share $\Delta$ with $h,m$\;
    \WHILE{($Left \neq \emptyset$)}
			\STATE Select $x$ from $Left$\;
                \STATE Let $s$ be a new session identifier\;
                \STATE $D := D \cup \{(s,x)\}$\;
			\STATE $j := 1$\;
                \STATE $l_m = l_h = \Init$\;  // dummy assignment
			\STATE $Done :=(j > n)$\;
			\WHILE{($\neg Done$)}
                    \STATE $\mu_m = (l_m,y_m,e_m):= \CallAgent(q_m,m,s,j,k,\Delta)$\; 
                    \STATE $M:= M \cup \{(s,j,m,\mu_m,h)\}$\;
                    \STATE $j := j + 1$\;
                    \STATE $Stop := ((l_m = \Ratify \wedge l_h = \Ratify) \vee (l_m = \Reject))$\;
                    \STATE $Done := ((j > n) \vee Stop)$\;
                    \IF{$\neg Done$}
				    \STATE $\mu_h = (l_h,y_h,e_h) := \CallAgent(q_h,h,s,j,k,\Delta)$\;
                        \STATE $M := M \cup \{(s,j,h,\mu_h,m)\}$\;
                        \STATE $j := j + 1$\;
                        \STATE $Stop := ((l_m = \Ratify \wedge l_h = \Ratify) \vee (l_h = \Reject$))\;
                        \STATE $Done := ((j > n) \vee Stop)$\;
                    \ENDIF
			\ENDWHILE
			\STATE $Left := Left \setminus \{x\}$\;
        \ENDWHILE
    \STATE \textbf{return} $\Delta$
    \end{algorithmic}
\end{algorithm}
In this section, we examine Human-LLM interaction through the use of the
\INT procedure. Experiments reported here look at two real-world problems:
{\textbf{X-Ray Diagnosis (RAD)}}, and 
{\textbf{Molecule Synthesis (DRUG).}} 
In the former, we want to use LLMs for diagnosing X-rays and producing reports, and
in the latter, we want proposals for synthesis pathways for molecules. We report on two kinds of experiments. First, \textit{controlled} experiments
are conducted using a database of human-authored predictions and explanations
and an LLM. This allows us to perform repeated (simulated)
experiments, which are needed since sampling variations can arise with the use of LLMs. Secondly, we report on results obtained in
\textit{uncontrolled} experiments using 
human subjects with varying levels of expertise interacting with an LLM.
% Controlled experiments are conducted on both RAD and DRUG; and
% uncontrolled experiments are reported on DRUG only.
We use the experiments to assess the following:
{\textbf{Human- and Machine-Intelligibility.}}
    We estimate the proportion of interactions
    exhibiting:(a) one- and two-way intelligibility; and(b) Strong- and Ultra-Strong Intelligibility.
{\textbf{Machine-Performance.}} We estimate the changes in machine performance 
    to increase when the interaction is at least one-way intelligible
    for the agent.\footnote{
    Human-Performance cannot be assessed with controlled experiments,
    as the human proxy is a database that is not updated. Human-Performance
    \textit{can} be assessed in uncontrolled experiments
    but it is difficult to establish baselines for comparative
    changes between specialists in complex domains
    with differing expertise.
    }
The results from uncontrolled experiments allow us to examine evidence to show that the simulation results
are representative of real-life usage of the \PXP protocol. 

% The results from uncontrolled experiments allow us to examine if
% there is any evidence to show that the results obtained in simulation
% on Human-Intelligibility are representative of real-life usage of the
% protocol. 

% \subsection{Materials}

\label{sec:mat}

% \subsubsection{Problems}
\subsection{Problems}

The \textbf{RAD} problem is concerned with obtaining predictions and explanations for
up to 5 diseases from X-ray images.
In this paper, an LLM is used to generate diagnoses and reports, given 
image data. We use data from the Radiopedia database~\cite{Radiopaedia07}. The \textbf{DRUG} problem is concerned with obtaining predictions and explanations for
the synthesis of small molecules. 
% This is one stage in the process of computational
% drug design. 
Once potential molecules (leads) have been identified for inhibiting a
target protein, we are interested
in synthesising these molecules for testing on
biological samples. In this paper, the DRUG problem examines the use of LLMs
to propose plans for synthesis, and an explanation of why the plan is proposed. We use data from DrugBank \cite{wishart:drugbank2018}. Dataset details are provided in Appendix \ref{sec:B_data}, and excerpts of sessions demonstrating the protocol's use in Appendix \ref{sec:D_sessions}.

% \subsubsection{Agents}
\subsection{Agents}

For RAD, the ``human-agent'' is an agent that has access to the database of the
human-derived predictions and explanations. In addition, to obtain the message
tags we require the agent includes a $\Match$ function
for comparing predictions and a $\Agree$ function for comparing explanations.
We implement these as follows:
% \begin{itemize}
\textbf{$\Match$}: Check whether the prediction of the machine-agent (disease/pathway) matches with the ground truth. This is a simple equality check. %for the Yes/No labels for RAD, and the synthesis pathways for DRUG.
\textbf{$\Agree$}: Check whether the explanation produced by the machine-agent is consistent with the explanation provided by the human-compiled database. This involves whether the two explanations concur. This is done here by querying a
    second (``tester'') LLM with a prompt similar to
    ``Are these two reports/pathways consistent with each other?''.
We note that since the predictions and explanations
are from an immutable database, there is no possibility of the human-agent in RAD sending a $\Revise$ tag. 
For DRUG, the human-agent has access to a chemist (one of the authors of this paper) who assesses the predictions and
explanations from the LLM. We assume the chemist nominally employs $\Match$ and $\Agree$ functions, these are not implemented as computational procedures. The chemist directly provides the message-tag along with predictions and explanations.
The chemist's message can be tagged with any of the tags allowed
in \PXP. For RAD and DRUG, the machine-agent uses an LLM for
generating predictions and explanations, and the machine-agent uses a separate LLM is used to implement $\Agree$. The task of this second (``checker'') LLM is to determine if the explanation obtained from the human-agent is consistent with the explanation generated by the machine-agent.

\subsection{Method}
\label{sec:meth}

We consider two kinds of experimental settings. The purpose of
\textit{controlled} experiments is to be able to perform repetitions.
These are conducted by way of simulations with a human-proxy, using a
database of human-authored predictions and explanations. These
experiments are conducted for both RAD and DRUG. We
also conduct \textit{uncontrolled} experiments, using 2 human chemists, with different levels of
chemical expertise (consistent with  graduate- and doctoral-level
training), and with different level of computational expertise. For \textbf{all} experiments, our method below is straightforward:
\begin{enumerate}

\item For $r = 1 \ldots R$:
\begin{enumerate}

     \item Initiate the \INT procedure with the data available
        obtain a record $\Delta$ on termination of the
        \INT procedure
    \item Store $\Delta$ as $\Delta_r$
 \end{enumerate}  
 \item Using the records in $\Delta_1,\ldots, \Delta_R$,
    obtain the frequency of sessions that are:
            (a) one-way and two-way Intelligible; and
            (b) Strong and Ultra-Strong Intelligible
\item Estimate the proportions of interest from the median
    values obtained above
    
\end{enumerate}
% We provide the following note of caution regarding all experiments.
The controlled
experiments consist of 5 repetitions with 20 instances (X-ray images for RAD,
and small molecules for DRUG). Uncontrolled experimental results are provided
with DRUG (radiologists were unavailable). 
% It is evident that sample sizes
% are small and we will focus on the main trends, rather than actual numbers.
 % The following specific details are relevant: 
 In \cite{pxp}, it is assumed that agents employ a \texttt{LEARN}
    function to decide whether to revise their predictions, based on estimating if revision would improve performance. In controlled experiments, we incorporate this through the use of a validation-set. Thus, the LLM only revises its prediction if its validation-set accuracy improves. In contrast, the human-proxy  functions as an oracle (predictions and
    explanations are always taken to be correct). In uncontrolled
    experiments, no such set is used by either agent. The \INT procedure requires a bound on the total messages exchanged, and is set to $10$. 
    % \textbf{(v)} As specified in the \PXP protocol, a session between human and machine
    %     exhibits one-way intelligibility
    %     for the human-agent if there is at least one
    %     entry in the message-table in a $\Delta$ with
    %     the human (machine) as sender that has a $\Revise$ or $\Ratify$ tag. A session
    %     is two-way Intelligible if it is one-way Intelligible for the human and one-way
    %     Intelligible for the machine;
% \textbf{(vi)} As proposed in~\cite{pxp},  a session between human and machine
%         is Strongly Intelligibility for the human-agent
%         if every entry in the message-table in a $\Delta$ with
%         the human (machine) as sender has a $\Revise$ or $\Ratify$ tag. A session
%         is Ultra-Strong for the human (machine) if it is Strongly Intelligible
%         and contains at least one entry in the message-table in $\Delta$ with
%         the human (machine) as sender that has a $\Revise$ tag.
We will use the usual (maximum-likelihood)
        estimate of proportion as the ratio of the number of sessions with a property
        to the total number of sessions;
We use Claude 3.5 Sonnet as the LLM, with more details in Appendix \ref{sec:A_expts}.

\subsection{Results}
\label{sec:results}

\begin{figure}[!htb]
    \centering
    \subfloat[\centering RAD]{{\includegraphics[width=0.33\textwidth]{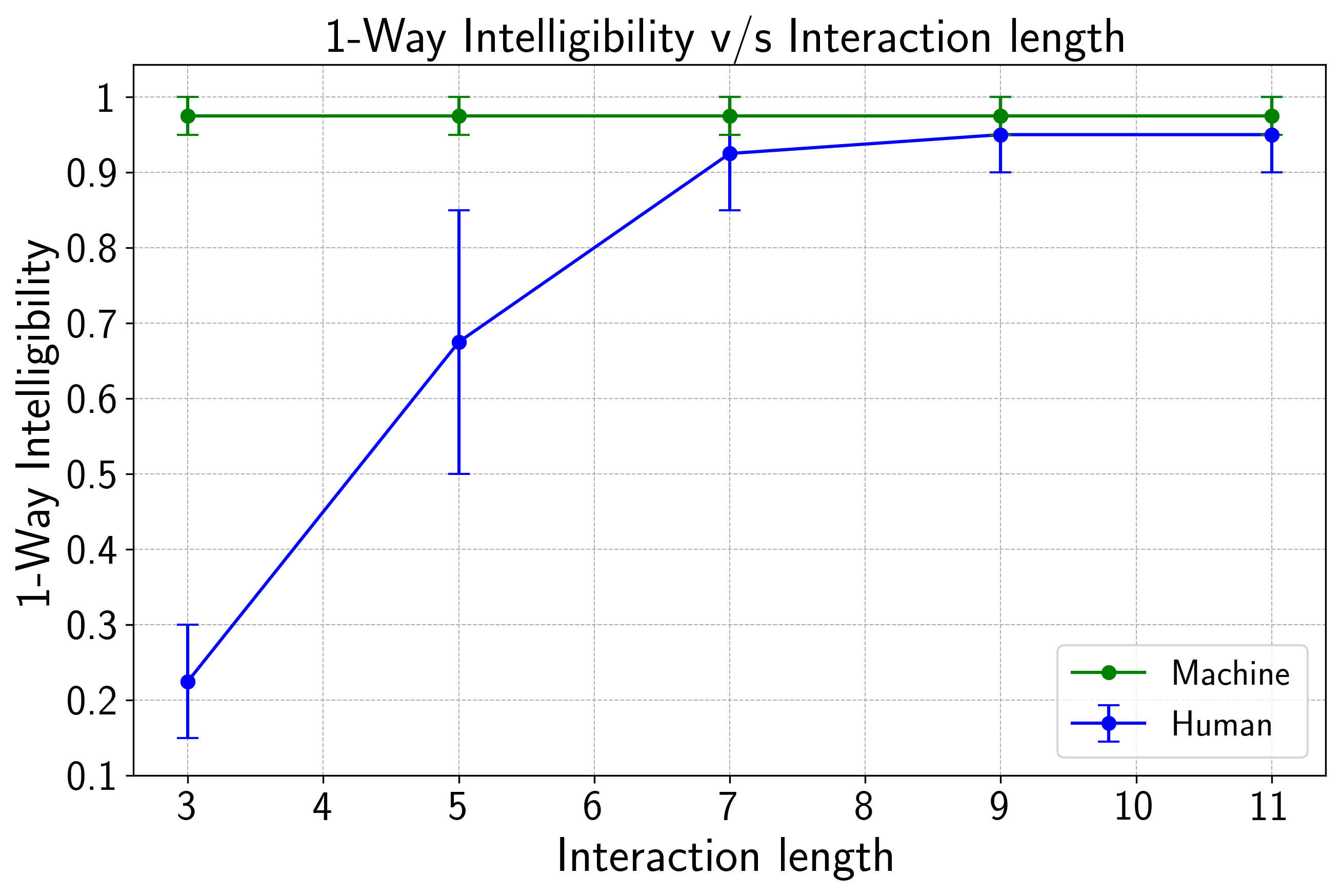} }}%
    % \qquad
    \subfloat[\centering DRUG]{{\includegraphics[width=0.33\textwidth]{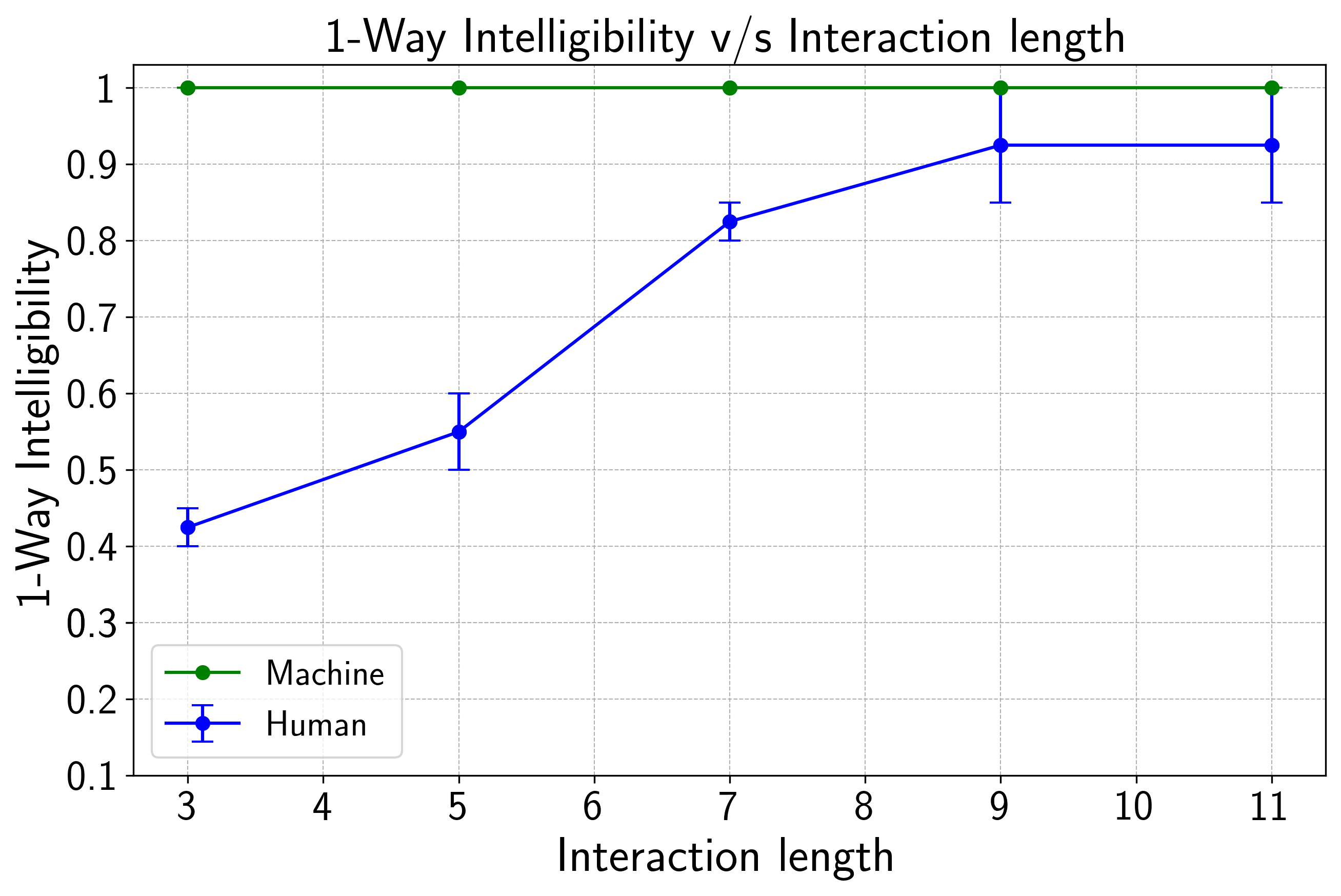} }}%
    % \qquad
    \subfloat[\centering DRUG - uncontrolled]{{\includegraphics[width=0.33\textwidth]{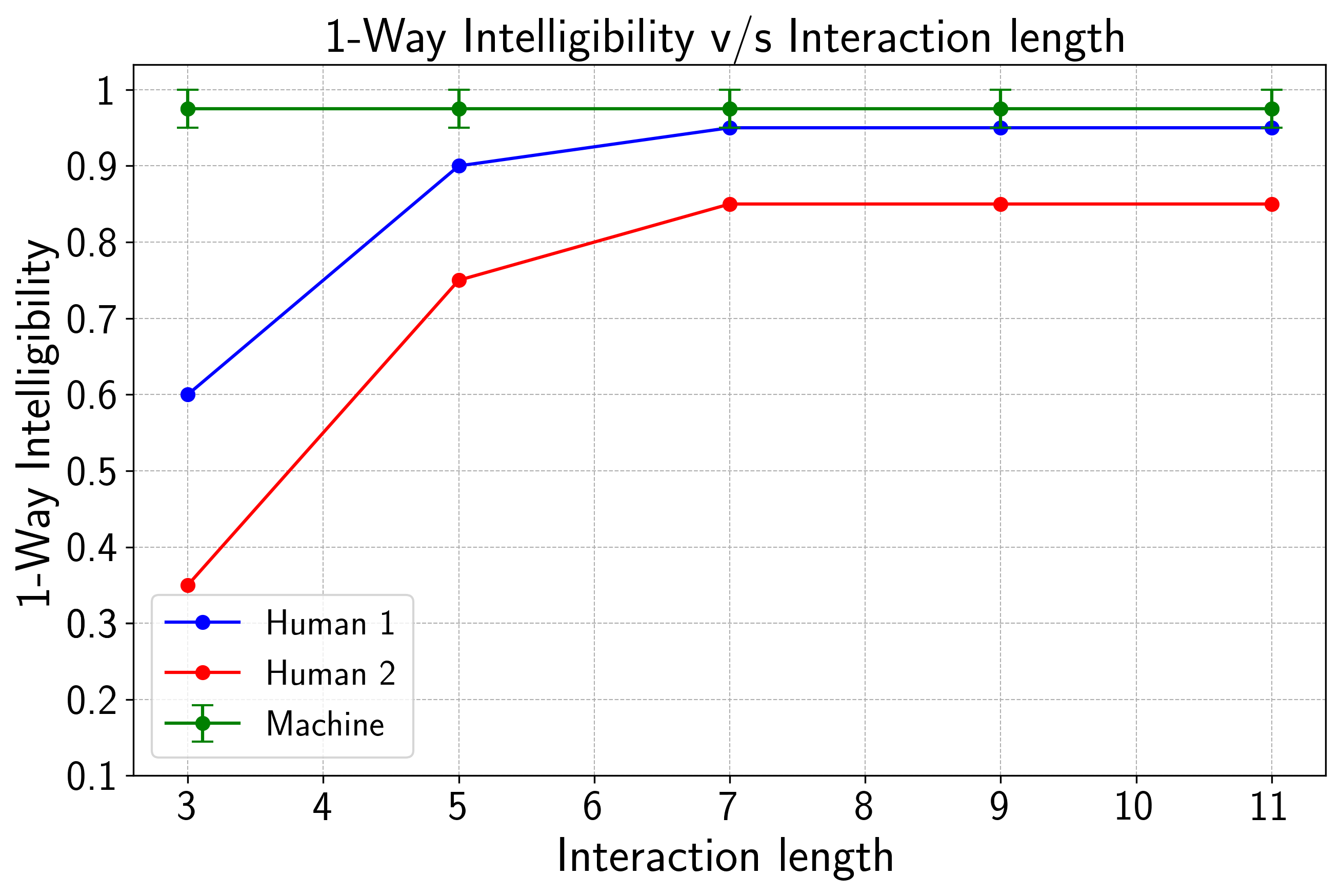} }}%
    \caption{Human- and Machine-Intelligibility in (a,b) controlled, and (c) uncontrolled experiments. The proportion of one-way intelligible sessions increases as the length of interaction is increased. In RAD, by message 3, 13 sessions are one-way intelligible for the human. Error bars are for 5 repetitions.}%
    \label{fig:results2}%
\end{figure}

\begin{figure}[!htb]
    \centering

    \subfloat[\centering Controlled]{{\includegraphics[width=0.33\textwidth]{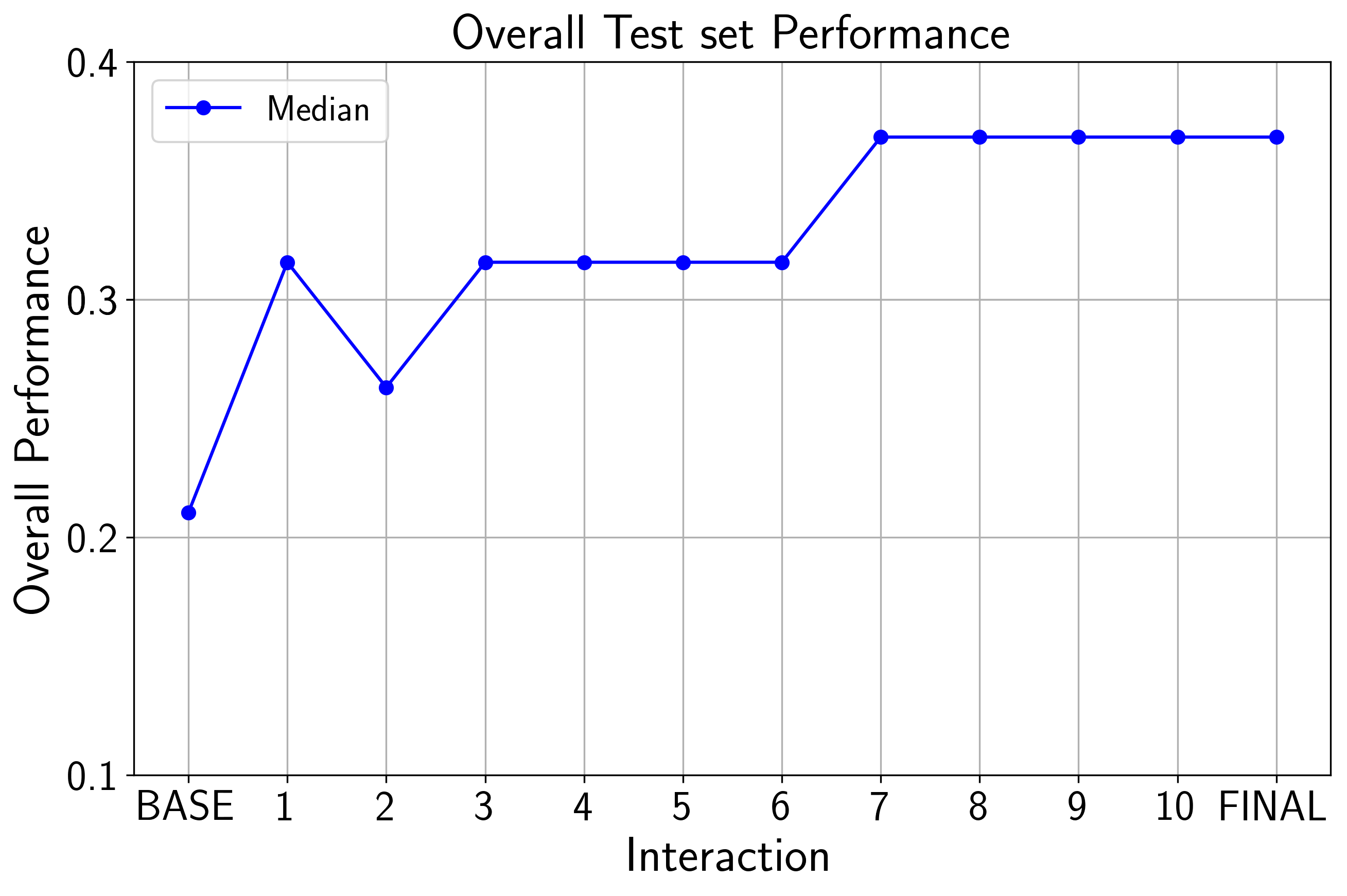} }}%
    \qquad
    \subfloat[\centering Uncontrolled]{{\includegraphics[width=0.33\textwidth]{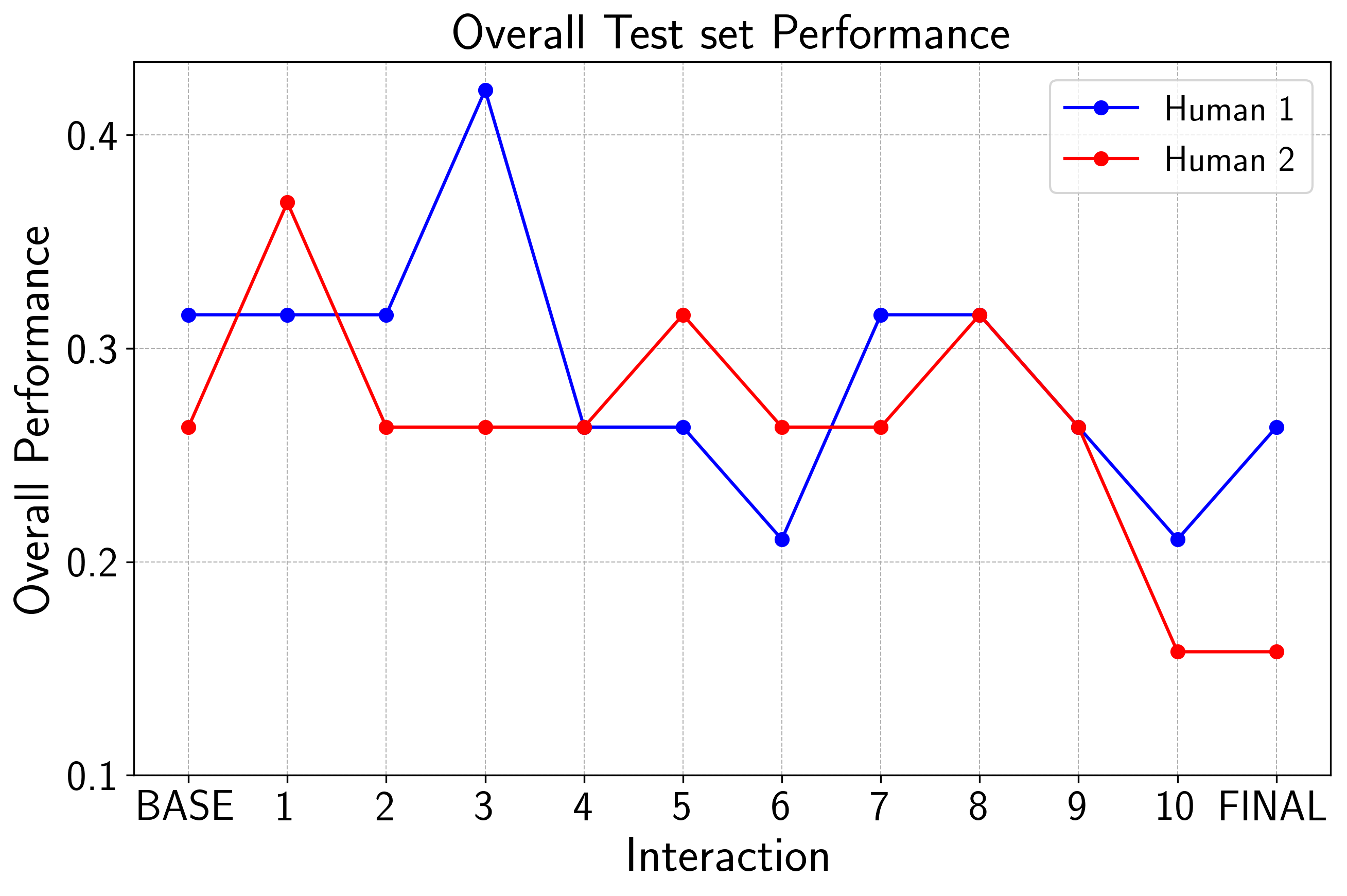} }}%
    % \qquad
    \caption{Machine-Performance in (a) controlled, (b) uncontrolled experiments.}%, test accuracy (Y) with interaction length (X).}%
    \label{fig:results3}%
\end{figure}

We focus first on the controlled experiments. For
simplicity, we will continue to call the human-proxy agent as ``the human agent''.
The statistics of interaction on controlled experiments are tabulated in
Tab.~\ref{tab:results1}, Fig.~\ref{fig:results2},~\ref{fig:results3}. Broadly, they are consistent
with the following claims:
\textbf{(a)} The proportion of one-way and
two-way intelligible sessions for the human agent
increases as the length of interaction increases.
\textbf{(b)} Machine-performance (measured on test data) increases with the length of interaction.
The following additional 
observations have possibly more interesting long-term
consequences of using the $\INT$ implementation of \PXP:
% \\
% \begin{description}
    \textbf{{One- and Two-Way Intelligibility:}} The frequencies
        of sessions that are 1-way intelligible are high for both human-
        and machine-agents in RAD, and very
        high in DRUG. Some of this is due
        simply to the vast store of prior data and information within an LLM
        that allows it to provide correct predictions and explanations almost
        immediately. For example, in 8 of 20 sessions in DRUG, the
        chemist and machine-agent agree on the prediction and explanations
        within 3 exchanges. However, interestingly, in
        an additional 7 sessions, they agree after some exchange of refutations
        and revisions. For human-in-the-loop systems,
        this is indicative of the advantage to the human of being provided with
        explanations in natural language. It is also evidence that the LLM is
        able to act on text-based feedback from the human, consistent
        with the findings in~\cite{llms:fewshot}. The number of
        sessions that are 2-way intelligible clearly cannot be more than the lower number of 1-way intelligible sessions. %\\
    \textbf{{Strong and Ultra-Strong Intelligibility}}: While it is not
        surprising to find the number of strongly intelligible sessions is
        lower than the number of 1- or 2-way intelligible sessions, the
        difference in numbers between strong- and ultra-strong sessions
        is surprisingly high. Resolving this requires examining first
        the sessions exhibiting strong intelligibility. Closer examination
        shows that these are exactly those sessions which are immediately
        ratified by both human- and machine-agents\footnote{For all of these
        the message-tag sequence is: $\langle \Init_m, \Ratify_h, \Ratify_m \rangle$.}.
        This is a degenerate form of strong-intelligibility (see Defn.~\ref{def:strong}).
        More interesting, strong-intelligibility would arise from longer sessions
        (for example, having a hypothetical sequence of message-tags:
        $\langle \Init_m, \Revise_h,$ $\Revise_m, \Ratify_h, \Ratify_m \rangle$).
        In fact, no such sequences occur. Thus, if we ignore these
        very short interactions, there are, in fact, very few strongly
        intelligible sessions. It is interesting though that in DRUG,
        we are able to observe 18 such sessions for the machine-agent.%\\
         \textbf{{Variability.}} The experiments were repeated 5 times.
        It is also evident from the error bars that sampling variation
        decreases as the length of sessions increases. We believe this
        to be due to the increasing context information being available to the
        generator LLM used by the machine-agent as session-length increases.

\begin{table}
    
    \caption{Interaction statistics. 
    % The counts are obtained by examining the message-tags sent in sessions bounded to a maximum of 10 messages (see Sec. \ref{sec:meth}). 
     For RAD, the median of 5 runs is reported. Only 1 run was possible for DRUG. Parentheses indicate proportions of the total sessions. In the uncontrolled experiments for DRUG, Human h1 has lower chemical but higher computational expertise than Human h2.}
    \label{tab:results1}
    \centering
    \small{
    \begin{tabular}{lrcccc}
    \toprule
    \multicolumn{2}{c}{Count}                             & RAD         & DRUG        & DRUG-h1     & DRUG-h2     \\
    \midrule
    \multicolumn{2}{l}{Total sessions}                    & 20          & 20          & 20          & 20          \\
    \midrule
    \multicolumn{1}{l}{1-way intelligible}       & Human   & 19 (0.95)   & 18 (0.90)   & 19 (0.95)   & 17 (0.85)   \\
    \multicolumn{1}{l}{sessions for:}            & Machine & 20 (1.00)   & 20 (1.00)   & 20 (1.00)   & 19 (0.95)   \\
    \midrule
    \multicolumn{2}{l}{2-way intelligible sessions:}      & 19 (0.95)   & 18 (0.90)   & 19 (0.95)   & 17 (0.85)   \\
    \midrule
    \multicolumn{1}{l}{Strong intelligible}       & Human   & 4 (0.20)    & 8 (0.40)    & 12 (0.60)   & 7 (0.35)    \\
    \multicolumn{1}{l}{sessions for:}             & Machine & 19 (0.95)   & 19 (0.95)   & 20 (1.00)   & 18 (0.90)   \\
    \midrule
    \multicolumn{1}{l}{Ultra-Strong intelligible} & Human   & 0 (0.00)    & 0 (0.00)    & 0 (0.00)    & 0 (0.00)    \\
    \multicolumn{1}{l}{sessions for:}            & Machine & 15 (0.75)   & 10 (0.50)   & 8 (0.40)    & 11 (0.55)   \\
    \bottomrule
    \end{tabular}
    }
\end{table}

% \noindent 
We turn now to results from the uncontrolled experiments, shown
in Fig.~\ref{fig:results3}(b). Broadly, we see the main trend
obtained in controlled experiments repeated here:
The proportion of intelligible sessions. It is interesting
however, to note that the human-agent with
higher computational--not chemical--expertise has a greater proportion
of intelligible sessions. Closer inspection showed this to be a chance
effect of the LLM's output (the same molecule can produce very different synthesis plans from the LLM). Thus, while the LLM's output remains
broadly intelligible, the level of intelligibility can vary from
one run to another.
The Machine-Performance plot also brings out an interesting aspect.
In the controlled experiments, the LLM's prediction was only
revised if performance on a validation-set improved. This check
is not done in the uncontrolled setting, thus violating
one of the assumptions of the protocol in \cite{pxp}. The result is that
if the assumption is violated, then we can end up in the paradoxical
position of the machine finding the human response intelligible, but that
not being reflected in improvements in performance.

\section{Conclusion}
\label{sec:concl}

Recently, we have seen a dramatic expansion in the use of ML, enabling it to support almost any activity where data can be collected and analysed. A challenge arises when the predictive power of modern ML meets the need for human understanding. On the face of it, the use of machine-agents that communicate in
natural language can mitigate the problem by providing explanations in a human-readable form.
But, readability, while necessary for understandability, may not be sufficient. In this
paper, we investigate the use of an information-exchange protocol specifically designed around
the notion of \textit{intelligibility} of messages exchanged between agents, that
attempts to address the broader issue of the quality of information exchanged in human–machine communication \cite{Coie:p:2001}. 

The concept of “intelligibility of communication” underpins the abstract interaction model proposed in \cite{pxp}, which explores how human–ML systems can collaborate to predict and explain data. While that work is conceptual, with no implementation, it presents case studies showing how an “intelligibility protocol” could qualitatively assess communication clarity between humans and machines.

This paper builds on that idea by implementing a simple interactive procedure that exchanges messages as proposed in \cite{pxp} and testing it on problems involving human interaction with large language models (LLMs). LLMs are well-suited for this because (1) their natural language capabilities enable effective collaboration with non-ML experts who have domain expertise, and (2) foundational LLMs possess extensive general knowledge useful for complex, data-driven analysis. The results are a first step toward
empirical support for using intelligibility as a foundation for designing collaborative
human–ML systems.

\newpage
\bibliographystyle{main}
\bibliography{main}

\appendix
\section{Experiment Details}
\label{sec:A_expts}

Here we detail the configuration for the LLMs used in the experiments.

All experiments use \texttt{claude-3-5-sonnet-latest} \cite{anthropic2024claude} as the LLM. For \texttt{ASK\_AGENT} (Algorithm \ref{alg:askagent}).
The \texttt{max\_tokens} for RAD is set to 300, and for DRUG it is set to 1024.
For \texttt{AGREE}, the \texttt{temperature} parameter for both RAD and DRUG is set to 0, and \texttt{max\_tokens} is set to 10. All other settings are the defaults, set by the Claude API.

% mention date of expts since `latest` is now outdated 

% \begin{itemize}
%     \item \textbf{(iii)}For RAD, the ``temperature'' parameter for the LLM used to check explanations
%         (used by both agents) is set to $0.0$. The LLM used to
%         to generate predictions and explanations by the machine-agent uses the
%         the default value of temperature in the OpenAI API ($1.0$);

%     \item \textbf{(iv)} For DRUG, the ``temperature'' parameter for the LLM used to check explanations by the machine is set to $0.0$. 
%     The LLM used to generate predictions and explanations by the machine-agent has temperature value of $0.3$;

%     \item \textbf{(viii)} The following LLMs were used. For both RAD and DRUG, the generation of predictions, explanations and the check for agreement in explanations was done by Claude-3.5 Sonnet~\cite{anthropic2024claude}.
% \end{itemize}

\section{Dataset Details}
\label{sec:B_data}

\textbf{RAD}:
We use Radiopaedia ~\cite{Radiopaedia07} for data.
This is a multi-modal dataset compiled and peer-reviewed by radiologists. Each
tuple in our database consists of: (a) An X-ray image; (b)
A prediction consisting of a set of diagnosed diseases, and (c) an explanation
in the form of a radiological report. Both (b) and (c) are from expert human
annotators. %An example of a database entry is shown in Fig.~\ref{fig:rad}.
We use a subset of the full Radiopedia database, focusing on 5 ailments
($Atelectasis$, $Pneumonia$, $Pneumothorax$, $Pleural\space Effusion$, and
$Cardiomegaly$) as our database
with 4 instances per disease (20 instances overall). We summarise the reports using \texttt{gpt-3.5-turbo-0125} and use these as the ground truth.

\textbf{DRUG}: 
Molecules are selected from DrugBank~\cite{wishart:drugbank2018} (drugs are
leads that satisfy additional biological and commercial constraints). 
It contains a list of drug molecules approved by the FDA (Food and Drug Administration).
Specifically, we will focus on a subset of molecules with a molecular weight between 150g/mol - 300g/mol and have at least one aromatic ring. This set has been further sampled to get 20 molecules.

\section{Message Tags}
\label{sec:C_tags}

The table below shows how a message is tagged in the \texttt{AGENT}  procedure (Algorithm \ref{alg:callagent}), in a simplified format.

\begin{figure}[h!]
\centering
\begin{tabular}{cc|c|c|c}
&\multicolumn{1}{c}{\mbox{}}&\multicolumn{2}{c}{Explanation} & \\
&\multicolumn{1}{c}{\mbox{}}&\multicolumn{1}{c}{ ${\mathtt{AGREE}}(e_n,e_m)$}&\multicolumn{1}{c}{$\neg {\mathtt{AGREE}}(e_n,e_m)$} & \\ \cline{3-4}
& ${\mathtt{MATCH}}(y_n,y_m)$ & $\Ratify$ & $\Refute$ or & \\
Pred. & & (A) & $\Revise$ (B) & \\ \cline{3-4}
&  $\neg {\mathtt{MATCH}}(y_n,y_m)$ & $\Refute$ or & $\Reject$ &  \\
& & $\Revise$ (C) & (D) & \\ \cline{3-4}
\end{tabular}
\caption{Message tag categories}
\label{fig:decision-matrix}
\end{figure}

\section{Example Sessions}
\label{sec:D_sessions}

\begin{figure}
    \centering
    \includegraphics[width=\linewidth]{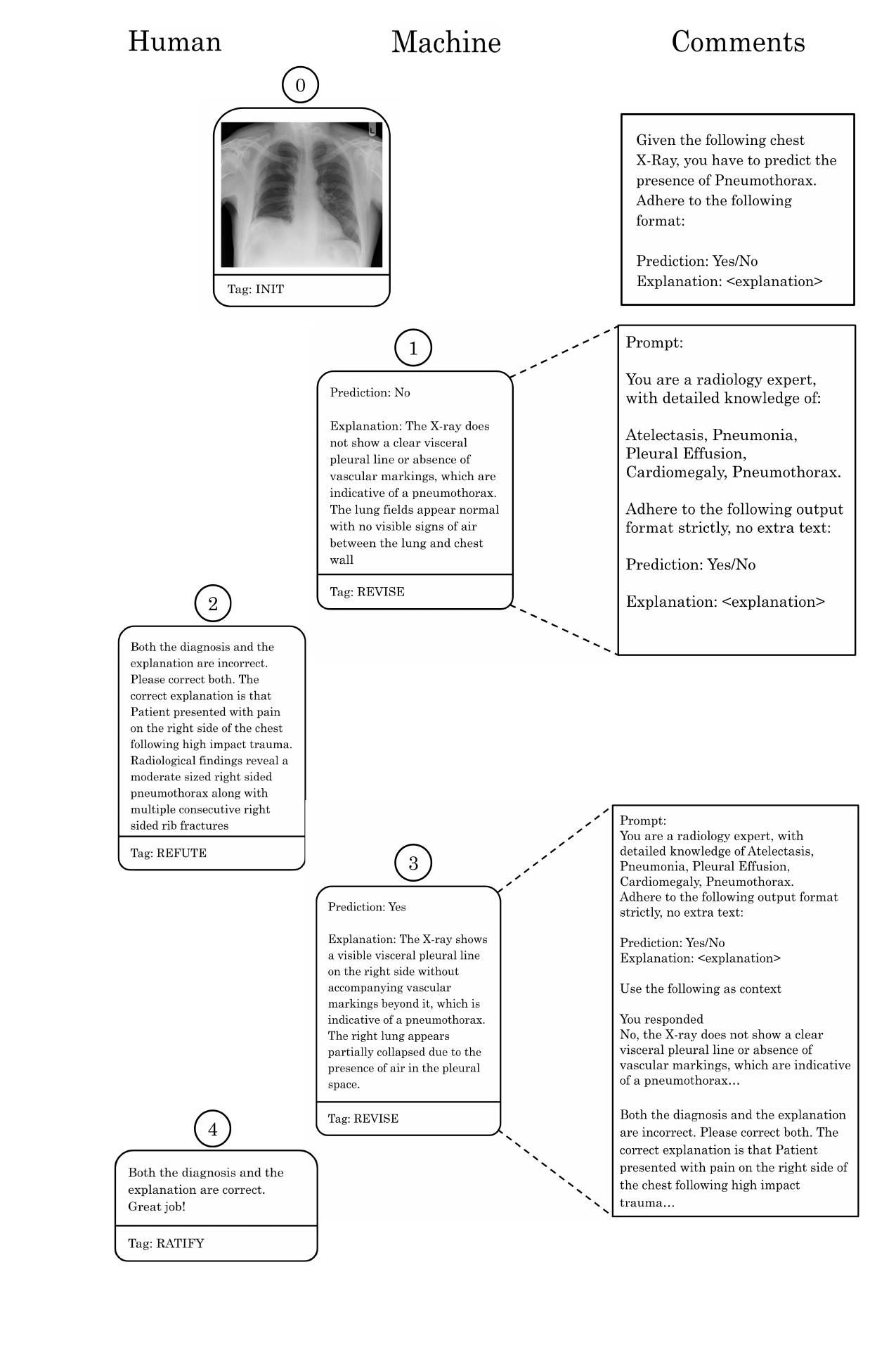}
    \caption{Excerpt of a session in the RAD experiment}
    \label{fig:conv_rad}
\end{figure}

\begin{figure}
    \centering
    \includegraphics[width=\linewidth]{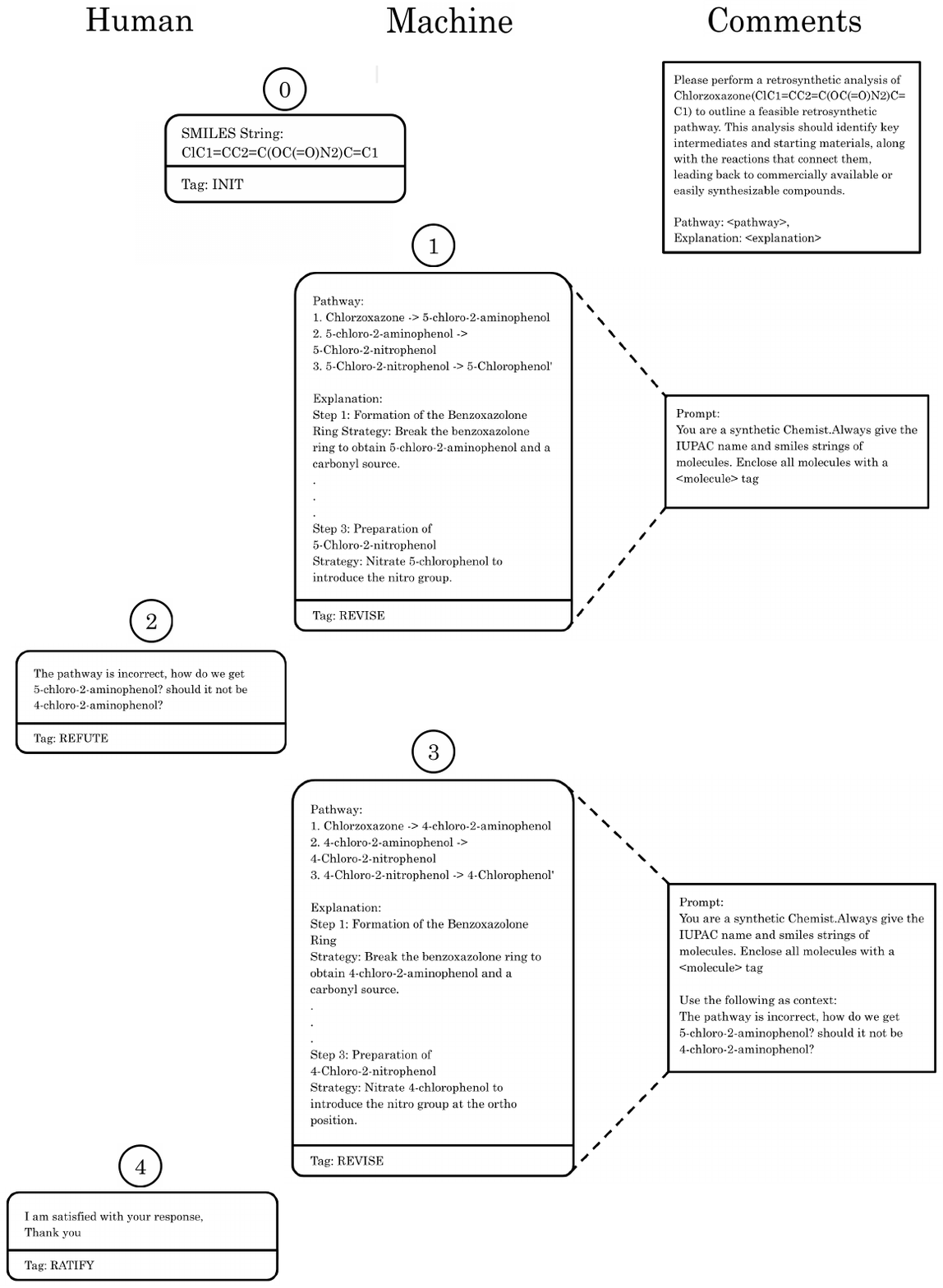}
    \caption{Excerpt of a session in the DRUG experiment}
    \label{fig:conv_drug}
\end{figure}

To illustrate the use of the $\Interact$ procedure (Alg. \ref{alg:hloop}), we show excerpts of the conversations betweeen agents for each of the tasks, Fig.\ref{fig:conv_rad} for RAD and Fig.\ref{fig:conv_drug} for DRUG. Note that the first input is provided to both the Human and Machine agent at the start with the $\Init$ tag, and is placed between the `Human' and `Machine' columns. The `Comments' column contains the prompts provided.

\end{document}